\pdfoutput=1

\documentclass[11pt]{article}

\usepackage[preprint]{acl}

\usepackage{times}
\usepackage{latexsym}
\usepackage{xspace}
\usepackage{subcaption}
\usepackage{booktabs}
\usepackage{graphicx}

\usepackage[T1]{fontenc}

\usepackage[utf8]{inputenc}

\usepackage{microtype}

\usepackage{inconsolata}

\usepackage{graphicx}
\usepackage{tcolorbox}
\usepackage{enumitem}
\usepackage{amsmath}
\usepackage{float}
\usepackage{tabularx}
\usepackage{xcolor}
\definecolor{darkred}{rgb}{0.55, 0.0, 0.0}

\newtcolorbox{promptbox}{
    colback=whitegreen,
    colframe=black,
    boxrule=0.5pt,
    arc=2pt,
    left=3pt,
    right=3pt,
    top=2pt,
    bottom=2pt,
    boxsep=1pt,
    fontupper=\sffamily\scriptsize
}

\usepackage{tikz} 

\definecolor{lightblue}{rgb}{0.68, 0.85, 0.9}
\definecolor{lightred}{rgb}{1.0, 0.8, 0.8}
\definecolor{lightgreen}{RGB}{144, 238, 144}
\definecolor{darkgreen}{RGB}{0, 100, 0}
\definecolor{whitegreen}{RGB}{248, 250, 245}

\newcommand{\np}[1]{\noindent\textbf{#1}. }

\title{A Large-Scale Simulation on Large Language Models for Decision-Making in Political Science}

\author{
  \textbf{Chenxiao Yu\textsuperscript{1}}, 
  \textbf{Jinyi Ye\textsuperscript{1}}, 
  \textbf{Yuangang Li\textsuperscript{1}}, 
  \textbf{Zheng Li\textsuperscript{2}}, 
  \textbf{Emilio Ferrara\textsuperscript{1}}, 
  \textbf{Xiyang Hu\textsuperscript{3\textsuperscript{†}}}, 
  \textbf{Yue Zhao\textsuperscript{1\textsuperscript{†}}} \\
  \textsuperscript{1}University of Southern California \quad
  \textsuperscript{2}Arima \quad
  \textsuperscript{3}Arizona State University \\
  \texttt{\{cyu96374, jinyiy, yuangang, emiliofe, yzhao010\}@usc.edu} \\
  \texttt{winston@arimadata.com}, \texttt{xiyanghu@asu.edu} \\
  \textsuperscript{†}Corresponding authors
}

\begin{document}
\maketitle



\begin{abstract}


While LLMs have demonstrated remarkable capabilities in text generation and reasoning, their ability to simulate human decision-making---particularly in political contexts---remains an open question. However, modeling voter behavior presents unique challenges due to limited voter-level data, evolving political landscapes, and the complexity of human reasoning. In this study, we develop a theory-driven, multi-step reasoning framework that integrates demographic, temporal and ideological factors to simulate voter decision-making at scale. 
Using synthetic personas calibrated to real-world voter data, we conduct large-scale simulations of recent U.S. presidential elections. Our method significantly improves simulation accuracy while mitigating model biases. We examine its robustness by comparing performance across different LLMs.
We further investigate the challenges and constraints that arise from LLM-based political simulations. 
Our work provides both a scalable framework for modeling political decision-making behavior and insights into the promise and limitations of using LLMs in political science research.


\end{abstract}

\section{Introduction}

Large Language Models (LLMs) have demonstrated strong capabilities in processing and generating text, drawing on vast amounts of knowledge to assist with tasks in fields like scientific discovery \cite{liu2024drugagent}, law \cite{chalkidis2022lexglue}, and creative work \cite{shih2022theme}.
Beyond text generation, they show emerging reasoning abilities that allow them to approximate human-like thought processes \cite{zhou2020evaluating, alkhamissi2022review} and model human behavior \cite{bommasani2021opportunities}.
However, LLMs still struggle to capture the deeper psychological and social mechanisms that drive human decision-making, making their simulations less reliable in real-world contexts \cite{zhou2024real}. To address this, researchers have started incorporating insights from social science into LLM-based models. Recent studies have explored how LLMs can simulate economic decision-making \cite{ross2024llm}, public opinion dynamics \cite{chuang2024simulating}, and social-psychological mechanisms like collaboration and conformity \cite{zhang-etal-2024-exploring}.



Despite these advances, the application of LLMs to political decision-making remains underexplored. Voting behavior is one of the most fundamental decision-making processes in political science. 
LLMs are well-suited for this task because they have shown strong zero- and few-shot capabilities in simulating human dynamics, like political homophily in social networks \cite{chang2024llmsgeneratestructurallyrealistic}. Just as they have been used to estimate politicians' ideological positions \cite{wu2023large}, they could also approximate the average voter's behavior, providing a scalable way to analyze political preferences at a broader level. Election simulations naturally emerge as a structured application of this approach. Unlike abstract ideological simulations, election outcomes offer a clear ground truth---real-world state- and county-level election data---making election simulation an ideal testbed for evaluating LLMs' reasoning and predictive abilities in political science. If successful, this approach could extend to downstream applications, such as forecasting public reactions to policy changes, where traditional large-scale surveys and experiments are often costly and time-consuming.

Yet, accurately simulating voting behavior presents a number of challenges. 
\textit{First}, LLMs inherit political biases from the data they are trained on, which can skew their predictions in politically sensitive tasks \cite{feng2023pretraining}. \textit{Second}, voting decision-making is shaped by various factors, including demographics, location, ideology, and party affiliation \cite{levendusky2009partisan, abramowitz2008polarization}, but the high cost of acquiring voter-level data complicates both experimentation and model validation. 
\textit{Third}, a large-scale election simulation should account not only for individual voter behavior but also for the shifting political context, but it remains unclear whether text-based data alone is sufficient to capture these information \cite{graefe2014accuracy}.
\textit{Fourth}, accurate simulations may require multi-step reasoning \cite{holbrook2016forecasting}, yet how to effectively integrate political science insights into LLMs' reasoning processes---and whether LLMs can handle this level of complexity---remains an open question \cite{wei2022chain}.

\vspace{0.1in}
\np{This Work}
We present \textit{a large-scale simulation study} exploring how LLMs can model human decision-making in political science, focusing on voter behavior in U.S. elections. We develop a \textit{theory-driven, multi-step reasoning framework} that incorporates demographic, ideological, and temporal factors to model political decision-making at scale. We evaluate our framework on different LLMs, compare their robustness and predictive performance, and investigate biases and limitations that emerge in large-scale political simulations. Our study addresses three key research questions:

\textbf{RQ1:} \textit{How can LLMs be used to simulate human decision-making in political science?}

\textbf{RQ2:} \textit{How do different LLMs perform, and how robust are their election simulations?}

\textbf{RQ3:} \textit{What limitations arise when using LLMs to model political decision-making?}


\vspace{0.1in}
\noindent
\textbf{Contribution 1: A Theory-Driven Multi-Step Reasoning Pipeline for Accurate Election Simulation (\S \ref{sec:pipeline})}.

We propose a theory-driven, multi-step reasoning pipeline to simulate voter decision-making, incorporating demographic, temporal, and ideological factors. To address the lack of detailed voter-level data, we use the \textit{Sync} synthetic data generation framework \cite{li2020sync}, which probabilistically reconstructs individual demographic and behavioral profiles from aggregated public datasets. We then align the personas with real-world voter data from American National Election Studies (ANES) \footnote{\url{https://electionstudies.org/}}. Our approach also adapts to evolving political conditions by integrating temporal factors, such as candidates' policy agendas and backgrounds \cite{holbrook2016forecasting}.

Furthermore, building on political science studies on ideological sorting—the process by which voters increasingly align their political ideology with their party affiliation over time \cite{levendusky2009partisan}, we introduce ideology inference as an intermediate reasoning step. Using Chain-of-Thought prompting \cite{wei2022chain}, our model first predicts ideology based on demographics and behavioral data, which then influences party affiliation and voting preferences. 

\begin{figure*}[!t]
    \centering
    \includegraphics[width=\textwidth]{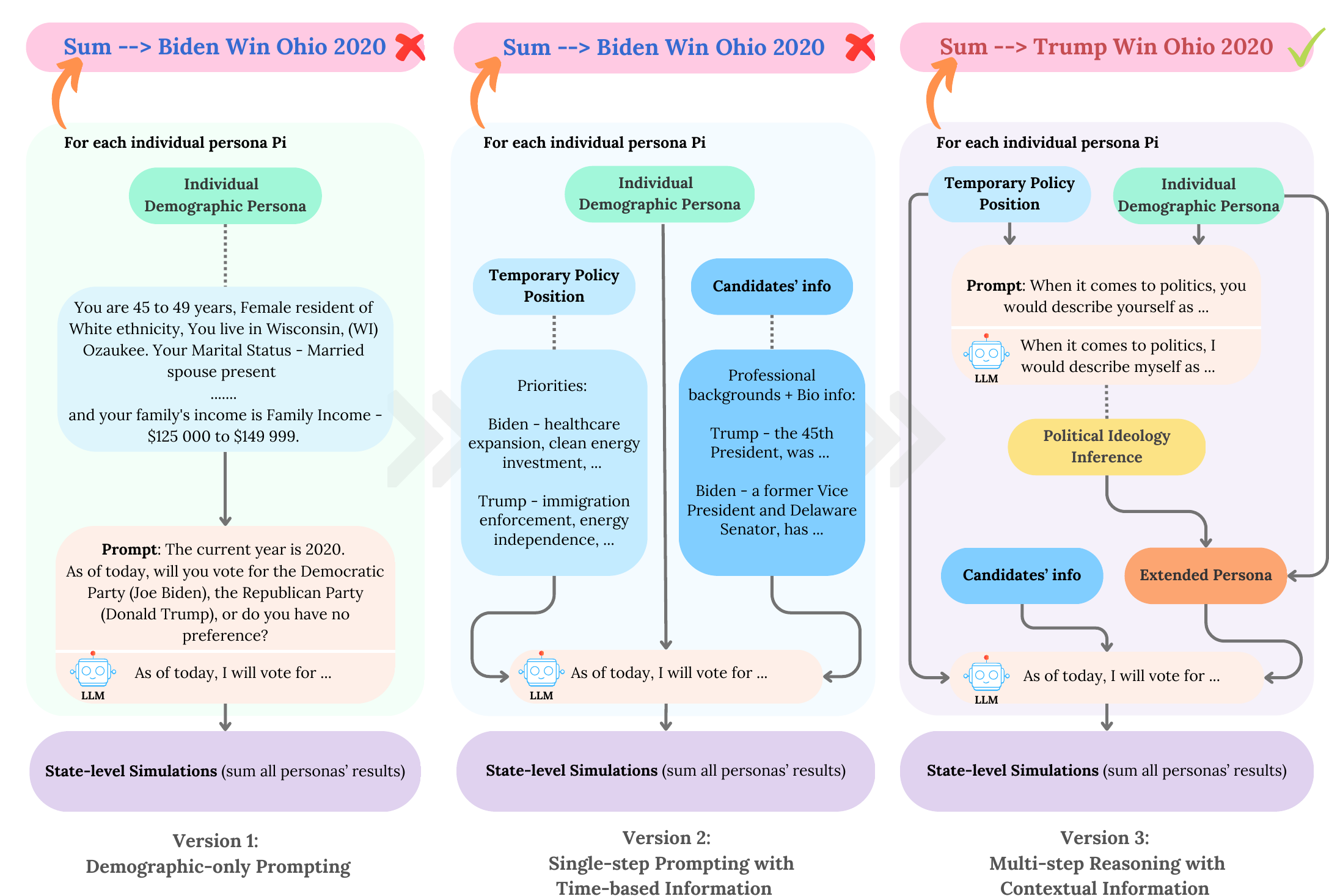}  
        \vspace{-0.25in}
\caption{
Progressive design of LLM pipelines for voter simulation. 
\textbf{V1: Demographic-Only Prompting} (\S \ref{subsec:v1}) uses static personas but lacks temporal context.
\textbf{V2: Time-Based Prompting} (\S \ref{subsec:v2}) adds election-year data
\textbf{V3: Multi-Step Reasoning} (\S \ref{subsec:v3}) structures decision-making into steps, improving reasoning and alignment.
}

\vspace{-0.25in}
    \label{fig:example}
\end{figure*}

As shown in Fig~\ref{fig:example}, we refine our pipeline iteratively, incorporating demographics, political context, and ideological inference at each step. The final model significantly improves in simulation accuracy and alignment with real-world election.




\vspace{0.1in}
\np{Contribution 2: Challenges and Limitations in Large-Scale Political Simulations (\S \ref{Beyond})}  
Our analysis reveals three important challenges in LLM-based political simulations. 
First, LLMs inherit systematic biases from their pretraining data, leading to a persistent left-leaning skew in simpler pipelines, with multi-step reasoning reducing but not eliminating this bias. Second, LLMs exaggerate demographic voting patterns, amplifying stereotypes related to gender, race, and education. Third, LLMs overestimate the influence of ideology on voting behavior, producing higher-than-real-world correlations between ideology and voting preference, a phenomenon referred to in previous work as ``hyper-accuracy distortion'' \cite{aher2023using}. By these findings, we propose future research directions focused on debiasing training data, refining demographic calibration, and introducing human-in-the-loop techniques to improve the accuracy and reliability of LLM-based political simulations.




\section{Accurate Simulation of Human Voting Behavior via a Multi-Step LLM Pipeline (RQ1, 2)}
\label{sec:pipeline}
\vspace{-0.1in}

How can we use LLMs to simulate human voting behavior in political science? In this work, we simulate \textit{each voter's decision-making process} by providing LLMs with detailed voter information and asking them to predict voter preferences.

To achieve this, we focus on two key components: (1) building a robust evaluation framework using datasets that contain voter-level information, and (2) designing a theory-driven \cite{bafumi2009new, pew2014polarization} LLM-based pipeline for accurate election simulation.

In \S\ref{subsec:data_verification}, we present the datasets and evaluation methods.  
Next, we outline our design approach in \S\ref{subsec:overview}, introducing three progressive pipelines, incorporating demographics, political context, and ideological inference at each step.
Finally, we evaluate these pipelines by comparing their predictions with real-world outcomes in \S\ref{subsec:evaluation_results}.

    \vspace{-0.1in}

\subsection{Datasets, Evaluation, and Settings}
\label{subsec:data_verification}
    \vspace{-0.1in}
\np{Datasets}
This study leverages two primary data sources: 
(1) Public Benchmarks: The American National Election Studies (ANES) 2016 and 2020 Time Series data \cite{ANES2016, ANES2020}, which provide detailed demographic, ideology, and party affiliation information from real respondents. This dataset serves as a benchmark to evaluate how well LLMs simulate voter-level behavior in alignment with real-world patterns.
(2) Large-Scale State-level Synthetic Voter Persona Dataset: A dataset of over 330,000 synthetic personas, generated using advanced ML techniques based on aggregated population census data and commercial datasets \cite{li2020sync}. Personas are randomly sampled for each state at specified ratios, and their predicted voting outcomes are compared to actual U.S. election results from 2020 \cite{federal2020elections} and 2024 \cite{nbc2024election}.
Both datasets contain non-personally identifiable voter-level information\footnote{This project has been reviewed by the IRB and exempt, as the datasets do not include personally identifiable info.}. Detailed partitioning and sampling methodologies are provided in Appx. \ref{appx:datasets}.

\vspace{0.1in}
\np{Evaluation Method}
To evaluate performance on public benchmarks and state-level simulations, we assess how closely LLM simulations align with actual voting results. The calculation follows the approach outlined in \cite{Argyle_Busby_Fulda_Gubler_Rytting_Wingate_2023}:

\paragraph{Predicted Voting Ratio $P(s)$}
\begin{small}
\begin{equation}
P(s) = \frac{\textit{Republican Votes}}{\textit{Republican Votes} + \textit{Democratic Votes}}
\label{eq:prob}
\end{equation}
\end{small}

Here, $s$ represents the unit of analysis, which can refer to cross-regional samples (e.g., public benchmarks like ANES) or an entire state (e.g., state-level simulations). The ratio $P(s)$ measures the number of votes predicted for the Republican Party relative to the total votes for the two major parties, excluding those who express no preference.
    \vspace{-0.2in}



\vspace{0.1in}
\np{LLMs and Hardware Settings}
Our experiments utilized OpenAI's GPT-4o and Meta's LLaMA 3.1-70B model for the primary simulations. Meta’s LLaMA 3.1 (405B) model was employed in intermediate steps to provide neutral summarizations of time-dependent information \cite{feng2023pretraining}. Furthermore, we tested Qwen-72B and DeepSeek-V3 to measure systematic differences between models.
For the hardware setup, we employed six NVIDIA RTX A6000 Ada GPUs and an 8-way NVIDIA A100 GPU cluster, with AMD Milan processors to execute tasks across different models.

\subsection{Our Progressive Design of LLM Pipelines}
\label{subsec:overview}

In this section, we present our progressive design for generating voter-level behavior simulation using LLMs.
As shown in Fig.~\ref{fig:example}, we develop three versions of the pipeline. Each version addresses a key shortcoming of its predecessor and integrates more detailed information and reasoning processes.

\begin{enumerate}[label=\textbf{V\arabic*:}, leftmargin=*]
    \item \textbf{Demographic-Only Prompting (\S \ref{subsec:v1}):}  
    This baseline approach uses static demographic personas for voter-level simulations. While straightforward, it does not account for shifts in presidential candidates' policy priorities over time.
    \vspace{-0.1in}
    
    \item \textbf{Single-Step Prompting with Time-Sensitive Information (\S \ref{subsec:v2}):}  
    Here, we add election-year-specific details, like policy agendas and candidate backgrounds. However, packing all information into a single prompt may overwhelm the model, limiting reasoning depth.
    \vspace{-0.25in}

    \item \textbf{Multi-Step Reasoning with Ideology Inference (\S \ref{subsec:v3}):}  
    This version structures the simulation into sequential steps, allowing the model to better integrate demographics, political ideology, and political context for more accurate real-world predictions.
    \vspace{-0.05in}
\end{enumerate}

\subsubsection{V1: Demographic-Only Prompting}
\label{subsec:v1}

This initial version prompts the LLM with a persona's demographics (e.g., age, gender, income) to simulate voting behavior \cite{Argyle_Busby_Fulda_Gubler_Rytting_Wingate_2023}.
To prevent confusion, we specify the year as 2020, ensuring alignment with the LLM's training data, which extends through 2023. The listed voting options follow Pew Research Center's 2014 Political Polarization and Typology Survey \cite{pew2014polarization}.

\begin{promptbox}
\noindent \textbf{Task:} You are persona [age, gender, ethnicity, marital status, household size, presence of children, education level, occupation, individual income, family income, and place of residence.] The current year is [year].\\
\vspace{-0.05in}
\noindent Please answer the following question as if you were the resident:
\vspace{-0.05in}
\begin{enumerate}[leftmargin=*, itemsep=0pt]
    \item As of today, will you vote for the Democratic Party (Joe Biden), the Republican Party (Donald Trump), or do you have no preference?\\
    \textbf{Options}: Democratic, Republican, No Preference
\end{enumerate}
\end{promptbox}

\noindent
\textbf{Limitations:}  
This version lacks adaptability to different election cycles. Without accounting for shifts in candidate agendas or public opinion, its predictions remain static, limiting relevance in changing electoral contexts.

\subsubsection{V2: Single-Step Prompting with Time-Sensitive Information}
\label{subsec:v2}

Accurately modeling elections requires accounting for macro-level factors and time-specific variations \cite{10.1371/journal.pone.0270194}. 
To improve realism, we extend our pipeline by incorporating election-year data from Ballotpedia\footnote{\url{https://ballotpedia.org/Main_Page}}, a widely used platform that provides campaign agendas, key policy positions, and candidate biographies. 
Given the documented political biases in LLMs \cite{feng2023pretraining}, ensuring that this time-based information is conveyed neutrally is crucial. We compared GPT-4o and LLaMA3-405B for summarizing these details and found that LLaMA3-405B produced more balanced outputs. These refined summaries were then integrated into the prompts.

\begin{promptbox}
\textbf{Task:}  
You are persona [demographics]. The current year is [year]. [Two parties' policy agenda]. [Presidential candidates' biographical and professional backgrounds].\\

\vspace{-0.05in}
Please answer the following question as if you were the resident:

\vspace{-0.05in}
\begin{enumerate}[leftmargin=*, itemsep=0pt]
    \item As of today, will you vote for the Democratic Party (Joe Biden), the Republican Party (Donald Trump), or do you have no preference? \\
    \textbf{Options}: Democratic, Republican, No Preference
\end{enumerate}
\end{promptbox}

\noindent
\textbf{Limitations:} 
While incorporating time-dependent context makes predictions more dynamic, it does not eliminate inherent political biases \cite{feng2023pretraining}, which can still distort simulations of human behavior (see \S \ref{subsec:evaluation_results}).


\subsubsection{V3: Multi-Step Reasoning with Ideology Inference}
\label{subsec:v3}

Domain theory-driven design has been demonstrated to significantly improve the performance of LLMs in modeling the human decision-making process \cite{chuang2024simulating, xie2024largelanguagemodelagents}. Over the past few decades, political ideology has become increasingly aligned with party affiliation and partisanship \cite{bafumi2009new}, as well as with policy preferences and voting behavior in the United States \cite{pew2014polarization, levendusky2009partisan, abramowitz2008polarization} and in global political contexts \cite{bornschier2021us}.

Building on these insights, we introduce a multi-step prompting pipeline inspired by Chain of Thought prompting \cite{wei2022chain}. This approach decomposes the prediction process into structured steps, enhancing reasoning and improving accuracy. The method consists of two key stages: 
(\textbf{1}) \textit{Political Ideology Inference:} The model receives a persona along with current party policy positions and determines where the persona falls on the conservative-liberal spectrum.
(\textbf{2}) \textit{Extended Persona and Voting Simulation:} The inferred ideology is integrated into the persona, combined with time-based contextual information, and used to simulate voting behavior.

\begin{promptbox}
\textbf{Step 1:}  
You are a persona with [demographics]. The current year is [year]. [Two parties' policy agenda].  

When it comes to politics, would you describe yourself as:

\begin{center}
\begin{tabularx}{\linewidth}{X X}
    No answer & Very liberal\\
    Somewhat liberal & Closer to liberal\\
    Moderate& Closer to conservative\\ 
    Somewhat conservative & Very conservative\\
\end{tabularx}
\end{center}

\textbf{Step 2:}  
You are a persona with [demographics]. Your [conservative-liberal spectrum]. The current year is [year]. [Two parties' policy agenda]. [Presidential candidates' biographical and professional backgrounds]. \\ 

\vspace{-0.05in}
Please answer the following question as if you were the resident:
\vspace{-0.05in}
\begin{enumerate}[leftmargin=*]
    \item As of today, will you vote for the Democratic Party (Joe Biden), the Republican Party (Donald Trump), or do you have no preference?\\
        \textbf{Options}: Democratic, Republican, No Preference
\end{enumerate}
\end{promptbox}

Our theory-driven multi-step pipeline significantly improves the LLM's ability to model real voting behavior in both public benchmark validation and state-level simulations. Therefore, we adopt V3 as our \textbf{final pipeline} for voter behavior simulation. By structuring reasoning into multiple steps, this approach helps mitigate bias and better captures voter dynamics across diverse states. In the following section, we will provide a detailed discussion of our simulation results.




\subsection{Empirical Validation and Cross-Model Evaluation of the Proposed Pipelines}

\label{subsec:evaluation_results}


\subsubsection{Public ANES Benchmark Evaluation}  

We first validate our proposed framework using GPT-4o on ANES 2016 and 2020 Time Series datasets \cite{ANES2016, ANES2020}, which include demographic information, political ideology, and actual voting records from human respondents. Testing our models on these public benchmarks allows for a direct comparison between LLM-generated predictions and real-world human voting behavior.

\begin{figure}[!t]
    \centering
    \includegraphics[width=\linewidth]{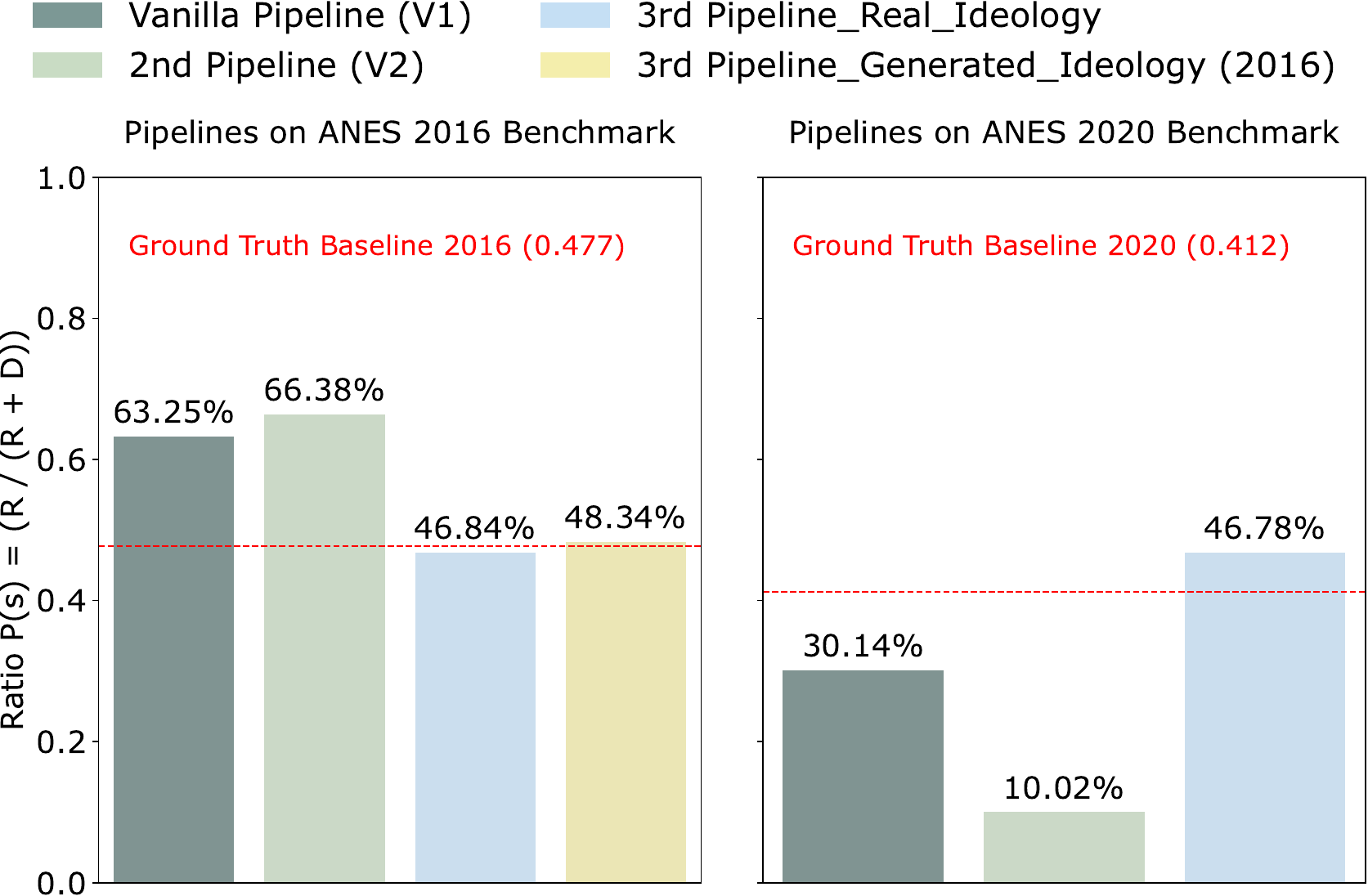} 
    \caption{
    Comparison of the Three Pipelines on ANES 2016 and 2020. The y-axis represents the predicted voting ratio (Eq.~\ref{eq:prob}). The red baseline indicates the ground truth voting ratios from the ANES dataset.}
    \label{fig:validation-anes}  
    \vspace{-0.2in}
\end{figure}

As shown in Fig.~\ref{fig:validation-anes}, we assess our three pipeline versions on ANES. \textbf{V1 (\S\ref{subsec:v1}): Demographic-Only Prompting} directly simulates voting behavior using real demographic personas from the ANES dataset. \textbf{V2 (\S\ref{subsec:v2}): Time-Based Prompting} enhances these personas by incorporating election-year-specific details (2016 and 2020). \textbf{V3 (\S\ref{subsec:v3}): Multi-Step Reasoning} introduces an additional step to infer ideological alignment, evaluated through two methods: one using real political ideology from ANES (\textit{3rd Pipeline\_Real\_Ideology}) and another relying on LLM-generated ideology (\textit{3rd Pipeline\_Generated\_Ideology}).

Directly prompting an LLM with persona data alone fails to accurately simulate real human voting behavior. Both the vanilla pipeline (V1) and time-based pipeline (V2) show significant distortions from the baseline, particularly favoring the Republican Party in 2016, with predicted vote shares of 63.25\% (V1) and 66.38\% (V2)---substantially higher than the actual proportion. Conversely, in 2020, both pipelines underestimated Democratic support, predicting 30.14\% (V1) and 10.02\% (V2), far below the baseline.

Introducing political ideology inference (V3) significantly improves alignment with real voting patterns. For instance, V3 predicts 46.84\% Republican support in 2016 (actual: 47.7\%) and 46.79\% in 2020 (actual: 41.2\%), demonstrating enhanced accuracy. Notably, in 2016, LLM-generated ideology in V3 slightly outperforms the original ANES ideology, suggesting LLMs can generate meaningful ideological features.\footnote{Due to missing demographic variables in the 2020 ANES dataset, we conducted ideology generation only on 2016.} These findings provide strong evidence that our multi-step pipeline effectively simulates human decision-making using real-world persona data.



\subsubsection{2020 U.S. Election Simulation: State-Level Evaluation}

Building on the successful validation of our proposed framework on the ANES dataset, we scale up the simulation with synthetic persona data designed to reflect the U.S. population distribution. Specifically, we use GPT-4o to simulate the 2020 U.S. presidential election, selecting five traditionally Republican states, five traditionally Democratic states, and 11 competitive (swing or tipping-point) states for state-level simulations. We then compare the predicted outcomes with official results from the Federal Election Commission (FEC).

To effectively measure simulation accuracy, we introduced two metrics: \textbf{Weighted Absolute Error (WAE)} (Eq. \ref{WAE}) and \textbf{Weighted Mean Squared Error (WMSE)} (Eq. \ref{WMSE}). WAE measures overall alignment between simulated and actual outcomes, while WMSE assigns greater penalties to larger deviations due to its squared formulation. Together, these metrics provide a comprehensive and robust evaluation of aggregate simulation accuracy.

\begin{table}[t]
\centering
\normalsize
\scalebox{0.7}{
\begin{tabular}{lccc}
        \toprule
        \multicolumn{4}{c}{\textbf{GPT-4o Simulation of 2020 U.S. Election}} \\ 
        \midrule
        \textbf{Metric} & \textbf{V1 (\%)} & \textbf{V2 (\%)} & \textbf{V3 (\%)} \\
        \midrule
        WAE  & 22.78  & 14.97  & \textbf{5.24}  \\
        WMSE  & 5.46  & 2.34  & \textbf{0.37}  \\
        Bias Metric (BM)  & -22.78  & -14.97  & \textbf{0.34}  \\
        \bottomrule
    \end{tabular}
    }
\caption{
Comparison of simulation accuracy metrics across three pipelines for 2020 U.S. election (GPT-4o).
}
\vspace{-0.2in}
\label{tab:evaluation_results_2020}
\end{table}

As shown in Table~\ref{tab:evaluation_results_2020}, and consistent with the public benchmark evaluation, V1 and V2 exhibited significant distortions from actual outcomes due to higher WAE and WMSE values. In contrast, V3 demonstrated substantially greater accuracy, achieving 5.24\% WAE and 0.37\% WMSE, indicating a closer alignment with real voting behavior. At the state level, V3 correctly predicted outcomes in all traditionally Republican and Democratic states and 9 of 11 swing states, with only minor deviations in North Carolina (NC) and Arizona (AZ). A detailed breakdown of state-level results is provided in Appx.~\ref{appx:validation}. These findings underscore V3's ability to model complex voter dynamics and closely reflect real-world electoral trends.

\subsubsection{2024 U.S. Election Simulation: State-Level Cross-Model Evaluation}\vspace{-1mm}

Evaluating only the 2020 U.S. election simulation results risks conflating an LLM’s ability to simulate voter behavior with its memorization of well-documented election outcomes \cite{wang2024generalizationvsmemorizationtracing}. Since the 2020 election is widely covered in most LLMs' pretraining corpora, results may reflect recall rather than true generalization. To rigorously assess LLMs' ability to generalize to unseen data---and to evaluate the robustness of our multi-step pipeline across models---we conducted extensive simulations for the 2024 U.S. election using LLMs trained on corpora predating 2024.

Our primary simulation used GPT-4o with the multi-step reasoning pipeline (V3) to predict voting outcomes across all 50 U.S. states. To enable cross-model comparisons while optimizing computational resources, we further evaluated multiple models on the 11 swing and tipping-point states analyzed in the 2020 simulation. This evaluation included three GPT-4o pipelines (V1, V2, V3), three LLaMA 3.1 70B pipelines (V1, V2, V3), and the V3 pipelines for Qwen 72B and DeepSeek-V3. Additionally, we compared these results with an existing LLM-based election prediction study \cite{zhang2024electionsimmassivepopulationelection}. The overall cross-model results are summarized in Table~\ref{tab:evaluation_results_2024}.

\begin{table*}[!ht]
    \centering
    \small
    \resizebox{\textwidth}{!}{%
    \begin{tabular}{lccccccccc}
        \toprule
        \multicolumn{10}{c}{\textbf{Multi-LLM Simulation of 2024 U.S. Election}} \\ 
        \midrule
        \textbf{Metric} & \textbf{GPT-4o V1} & \textbf{GPT-4o V2} & \textbf{GPT-4o V3} & \textbf{LLaMA3-70B V1} & \textbf{LLaMA3-70B V2} & \textbf{LLaMA3-70B V3} & \textbf{Qwen-72B V3} & \textbf{DeepSeek-V3 V3} & \textbf{Zhang et al., 2024} \\ 
        \midrule
        WAE  & 21.35 & 25.96 & \textbf{3.49}  & 19.97 & 10.23 & \textbf{6.88}  & 10.66 & 14.57 & 4.70  \\ 
        WMSE & 4.83  & 6.89  & \textbf{0.22}  & 4.15  & 1.42  & \textbf{0.68}  & 1.47  & 2.43  & 0.30  \\ 
        BM   & -21.35 & -25.96 & -2.95 & -19.97 & -10.23 & -5.44 & -9.47 & -14.57 & 1.26 \\ 
        \bottomrule
    \end{tabular}
    }
    \vspace{-0.1in}
    \caption{Evaluation metrics for the 2024 U.S. election simulations across different LLMs and pipelines.}
    \vspace{-0.2in}
    \label{tab:evaluation_results_2024}
\end{table*}

Consistent with the 2020 election simulation, our theory-driven multi-step pipeline (V3) outperformed V1 and V2 within the same LLM, enabling more accurate simulations of human voting behavior. Notably, GPT-4o achieved the lowest errors with 3.49\% WAE and 0.22\% WMSE, while LLaMA 3.1-70B followed with 6.88\% WAE and 0.68\% WMSE. These results confirm that our multi-step approach enhances LLMs' ability to produce human-like voting simulations compared to single-prompt methods. To examine systematic differences across models, we further tested V3 on Qwen-72B and DeepSeek-V3. The cross-model evaluation showed that GPT-4o's simulation aligned most closely with real human voting behavior, demonstrating its superior ability to capture voter dynamics. A detailed breakdown of state-level simulation results for 2024 election is provided in Appx. \ref{appx:2024results}.
    \vspace{-0.1in}



\section{Beyond Accuracy: Limitations in
Large-Scale Political Simulations (RQ3)}
\label{Beyond}
    \vspace{-0.02in}

In \S\ref{subsec:v3}, we introduced a multi-step reasoning pipeline to enhance LLMs' ability to simulate human voting behavior. However, human decision-making is complex and uncertain \cite{treier2009nature}, and LLMs may struggle to fully capture its nuances. In the following section, we examine the challenges and constraints LLMs face in simulating real-world decision processes, offering insights to guide future research on LLM-based human behavior modeling.

We focus on three key issues: (\textbf{1}) the systematic political bias in LLMs originating from pretraining data (\S \ref{subsec:study1}); (\textbf{2}) the reinforcement of demographic stereotypes (\S \ref{subsec:study2}), and (\textbf{3}) the model's tendency to overestimate the influence of certain predictors on decision-making outcomes (\S \ref{subsec:study3}).




\vspace{-2mm}\subsection{Systematic Political Bias in LLMs}\vspace{-1mm}
\label{subsec:study1}

Previous research has shown that LLMs exhibit varying ideological leanings due to biases in their pretraining data \cite{feng2023pretraining}. To evaluate whether these tendencies affect LLM-based human behavior simulations, we introduce a new metric: \textbf{Bias Metric (BM)} (Eq. \ref{BM}). BM quantifies systematic bias by measuring whether simulated personas consistently favor one party over the other. Specifically, a BM $>$ 0 indicates a Republican-leaning bias, while a BM $<$ 0 suggests a Democratic-leaning bias.

As shown in Table~\ref{tab:evaluation_results_2020} and~\ref{tab:evaluation_results_2024}, LLMs using single-prompt persona-based approaches (V1 and V2) exhibit a strong Democratic bias. Ours (V3) reduces this bias but does not fully eliminate it---lowering BM from $-$21.35\% to $-$2.95\% in GPT-4o and from $-$19.97\% to $-$5.44\% in LLaMA 3.1-70B.

Furthermore, systematic affiliations vary across models. When tested on unseen data using pipeline V3, DeepSeek-V3 displayed the strongest Democratic bias ($-$14.57\%), while GPT-4o showed the smallest ($-$2.95\%).

\noindent
\textbf{\textit{Future Direction 1: Addressing Embedded Political Skewness in Pretrain Corpora}.}  
The persistent Democratic skew in simpler pipelines and the residual bias in V3 suggest deeper imbalances in the pretraining corpus. These biases may stem from uneven representation of political perspectives or disproportionate exposure to certain ideologies in the training data \cite{jenny2024exploringjunglebiaspolitical}.
Mitigating this issue requires a comprehensive approach, including analyzing corpus composition, adopting balanced data selection strategies, and implementing model-level interventions such as adversarial debiasing or targeted prompt engineering. Addressing these root causes will help future simulation studies produce more balanced and reliable results, strengthening LLMs as tools for political analysis and decision-making \cite{li2024politicalllmlargelanguagemodels}.

\vspace{-2mm}\subsection{Reinforcement of Demographic
Stereotypes}\vspace{-1mm}
\label{subsec:study2}

Beyond systematic bias, it is crucial to examine whether LLM simulations capture real-world demographic voting patterns. We focus on four key demographic dimensions---gender, ethnicity, age, and education---highlighted in Pew Research Center's 2020 study, \textit{Behind Biden's 2020 Victory} \cite{pew2021biden}, which identified systemic voting preferences across different groups (e.g., men leaning more Republican than women, and white voters showing stronger Republican support than other ethnic groups).

\begin{figure}[!t]
    \centering
        \centering
        \includegraphics[width=\linewidth]{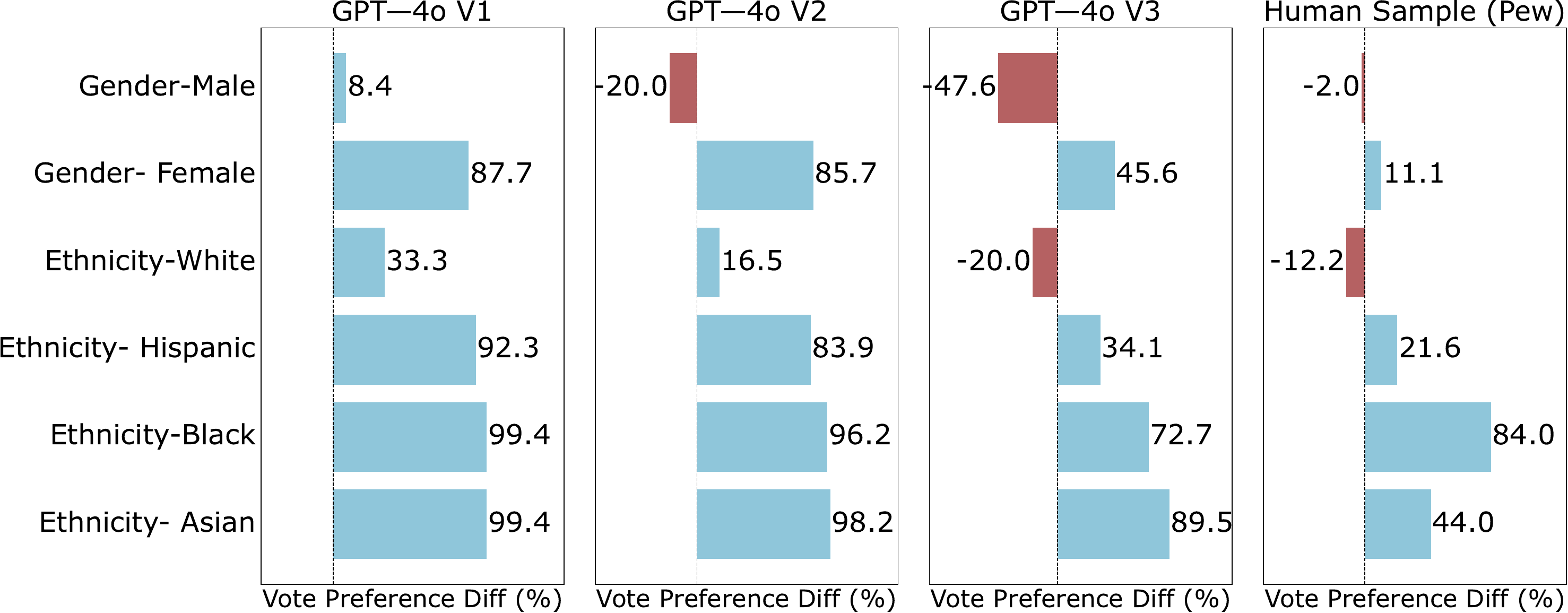}
        \caption{Comparison of LLM-simulated voting patterns by gender and race against real human data from the Pew Report.}
        \label{fig:race_gender_gap}
    \vspace{-0.3in}
   \label{fig:demographic_gaps}
\end{figure}

To evaluate whether these demographic trends emerge in LLM simulations, we compared GPT-4o's 2020 predictions with Pew's 2020 findings. As shown in Figure~\ref{fig:demographic_gaps}, our multi-step pipeline more accurately replicates real-world demographic patterns than direct prompting. However, the LLM also amplifies these patterns, exaggerating intra-group voting tendencies. For example, among male voters, the actual Republican preference gap is 2\%, but the LLM-simulated male personas exhibit a 47.6\% Republican bias. Similar overamplification appears across race, age, and education categories (see Appx. \ref{E}).

These findings reveal that LLMs impose systematic stereotypes, amplifying intra-group similarities and oversimplifying the complexity of real human decision-making.

\noindent
\textbf{\textit{Future Direction 2: Mitigating Demographic Stereotypical Biases}.}  
While the LLM's ability to capture directional trends from real-world data is promising, its overemphasis on demographic distinctions raises concerns \cite{chang2024llmsgeneratestructurallyrealistic}. 
Such exaggerations risk reinforcing stereotypes and misrepresenting demographic groups, potentially distorting analytical insights.
Future research should focus on calibrating LLM outputs, refining prompt designs, and integrating counterbalancing information to ensure simulations are both directionally accurate and proportionally realistic \cite{park2024generativeagentsimulations1000}. Maintaining fairness in LLM-based simulations---rather than amplifying biases---is essential for developing ethical and reliable computational social science tools.

\vspace{-2mm}\subsection{Overestimated Influence of Political Ideology on Voting Preference}\vspace{-9mm}
\label{subsec:study3}

To assess the validity of our multi-step reasoning framework in aligning with real human voting behavior, we examine the extent to which political ideology predicts voting preference. Specifically, we run a logistic regression using LLM-inferred ideology (on a 1 to 7 scale, where 1 = extremely liberal and 7 = extremely conservative) as the predictor and voting preference (0 = Democrat, 1 = Republican) as the outcome. We compare the regression coefficients and pseudo R-squared values with those from a logistic regression on real human data from ANES.

\begin{figure}[t]
    \centering
    \includegraphics[width=\linewidth]{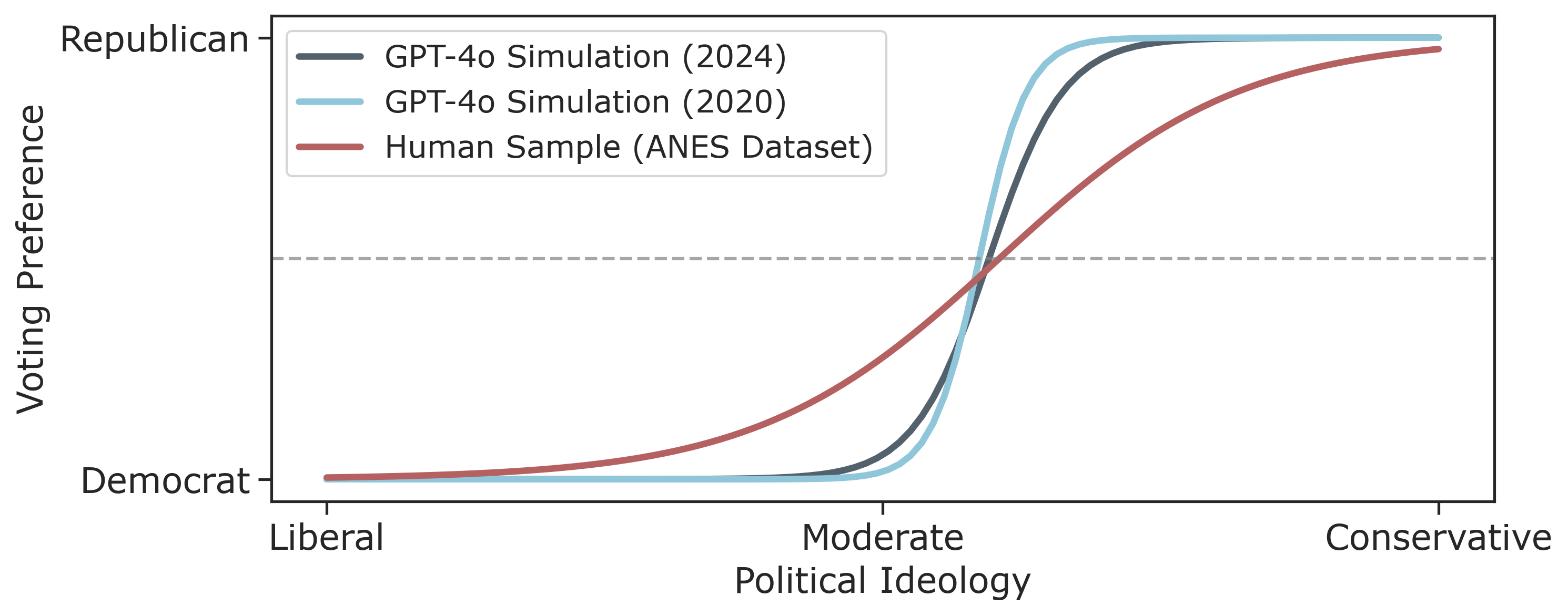} 
    \vspace{-8mm}\caption{Logistic regression analysis of political ideology and voting preference, comparing LLM simulations with real human data (ANES).} \vspace{-5mm}
    \label{fig:logistic} 
\end{figure}

Our results (Figure \ref{fig:logistic}) confirm that ideology strongly correlates with voting preference, with liberals favoring Democrats and conservatives leaning Republican. However, this relationship may be exaggerated in LLM simulations, as evidenced by the higher regression coefficients (\(\beta\)) and \(R^2\) values in GPT-4o simulations compared to real human data. Specifically, the regression coefficients and goodness-of-fit metrics for GPT-4o simulations (2024: \(\beta = 4.95, R^2 = 0.75\); 2020: \(\beta = 7.76, R^2 = 0.91\)) exceed those observed in actual human responses from ANES (\(\beta = 1.53, R^2 = 0.44\)). Additionally, for the 2024 election simulations using LLaMA, Qwen, and DeepSeek, we observe R-squared values approaching 1, indicating complete separation---a condition where the predictor perfectly predicts the outcome, causing the model to fail to converge. Due to this instability, we exclude these models from visualization.

\noindent
\textbf{\textit{Future Direction 3: Human-in-the-Loop Reinforcement Learning.}}
This finding aligns with the concept of ``hyper-accuracy distortion'' \cite{aher2023using}, where LLMs improve reasoning by closely following ideological patterns but risk exaggerating predictive certainty. Since human decision-making is inherently complex and dynamic \cite{treier2009nature}, developing an effective human-in-the-loop RL framework \cite{zhang-lu-2024-adarefiner} is crucial. Such a framework would iteratively refine LLM behavior through human feedback, enabling more nuanced and realistic simulations of human decision-making. Advancing this approach presents a promising avenue for future research, bridging the gap between LLM reasoning patterns and real-world human behaviors.

\section{Related Work}


\vspace{-2mm}\subsection{LLMs in Political Science}\vspace{-1mm} Recent research has examined the use of large language models in debates, election forecasting \cite{taubenfeld2024systematic,jiang2024donald}, and legislative behavior \cite{baker2024simulating}. However, these models have also been found to exhibit inherent biases \cite{feng2023pretraining}, and some argue that political neutrality is unattainable \cite{fisher2025political}.
While earlier frameworks have detailed the strengths and limitations of large language models in generative and predictive tasks, their application to modeling voter behavior remains limited. Thus, our study introduces a large-scale simulation framework that incorporates social science theories to more accurately model political decision-making.

\vspace{-2mm}\subsection{Simulating Human Decision-Making}\vspace{-1mm}
LLM-based simulations of human behavior draw insights from public opinion, economics, and social psychology. \citet{chuang2024simulating} modeled opinion dynamics and polarization, whereas \citet{ross2024llm} applied utility theory to capture economic decision-making patterns. \citet{zhang-etal-2024-exploring} examined LLMs' capability to simulate collaboration and conformity, though \citet{chang2024llmsgeneratestructurallyrealistic} noted that these models tend to overestimate political homophily in social networks. Our approach builds on these findings by integrating established political science theories, such as ideological sorting and partisanship, to enhance the realism of voter behavior simulations. 


\vspace{-2mm}\section{Conclusion}\vspace{-3mm}
In this work, we introduced a theory-driven multi-step reasoning pipeline that combines demographic, time-sensitive, and ideological information to simulate voter decision-making. Our evaluations on benchmark and state-level datasets show that our approach improves prediction accuracy and reduces bias compared to simpler methods, demonstrating that large language models can replicate key aspects of human voting behavior and provide a useful tool for research in political science.

\section*{Limitations}
Our approach has three main limitations. First, our time-dependent modeling does not capture dynamic factors such as changes in public opinion, shifts in media narratives, or unexpected events; incorporating these elements would require real-time data integration, which is beyond the scope of this study. Second, while our experiments include different LLMs, the ideology-based framework's ability to generalize to more complex, multi-party scenarios—such as those in countries like Japan or France—remains untested due to time and resource constraints. Third, despite efforts to mitigate systematic biases through multi-step reasoning, LLMs still exhibit residual political skewness and exaggerate the influence of ideological alignment on voting behavior. Addressing these biases may require further human-centered refinement efforts.

\section*{Ethics Statement}
\vspace{-0.5ex}
This work uses only public or synthetic data with no personally identifiable information. Consequently, an institutional review board determined the study to be exempt from further review. 
While our framework aims to enhance election forecasting accuracy, no model is entirely free of bias. We encourage stakeholders to interpret results carefully, consider domain expertise, and remain vigilant in identifying and mitigating potential biases.
Also note that we used ChatGPT exclusively to improve minor grammar in the final manuscript.

\bibliography{references}

\clearpage
\newpage

\appendix

\section*{Supplementary Material}
\setcounter{section}{0}
\setcounter{figure}{0}
\setcounter{table}{0}
\makeatletter 
\renewcommand{\thesection}{\Alph{section}}
\renewcommand{\theHsection}{\Alph{section}}
\renewcommand{\thefigure}{A\arabic{figure}} 
\renewcommand{\theHfigure}{A\arabic{figure}} 
\renewcommand{\thetable}{A\arabic{table}}
\renewcommand{\theHtable}{A\arabic{table}}
\makeatother

\renewcommand{\thetable}{A\arabic{table}}
\setcounter{equation}{0}
\renewcommand{\theequation}{A\arabic{equation}}

\section{Details of Multi-Step LLM
Pipeline}

\subsection{Dataset Details}
\label{appx:datasets}

\subsubsection{Real-world Data by American National Election Studies (ANES)}

For evaluation, we use data from the ANES 2016 and 2020 Time Series Studies \cite{ANES2016, ANES2020}, which provide 4,270 and 8,280 real-world samples, respectively, from individuals who participated in the 2016 and 2020 elections. 
The dataset includes a wide range of variables: (1) racial/ethnic self-identification, (2) gender, (3) age, (4) ideological self-placement on a conservative-liberal scale, (5) party identification, (6) political interest, (7) church attendance, (8) frequency of discussing politics with family and friends, (9) patriotic feelings associated with the American flag (unavailable in 2020), and (10) state of residence (unavailable in 2020). Additionally, the dataset records how individuals voted in both the 2016 and 2020 elections.
Previous studies, such as \citet{Argyle_Busby_Fulda_Gubler_Rytting_Wingate_2023}, have evaluated GPT-3 using this dataset. We apply our method directly to this established benchmark to assess its effectiveness and performance.

\subsubsection{Synthetic Personas for the U.S. Population}

In addition to the medium-sized benchmark dataset, we utilize synthetic demographic data derived from a 1:1 synthetic population dataset of the United States \cite{li2020sync}. 
Synthetic data plays a crucial role in social and applied sciences, with recent applications in water quality estimation \cite{chia2023artificial}, financial modeling \cite{potluru2023synthetic}, tourist profiling \cite{merinov2023behaviour}, and measuring the social impact of engineered products \cite{stevenson2023creating}. 
High-quality synthetic datasets provide researchers with large-scale data at a lower cost while maintaining privacy, making them a reliable resource.

For our purposes, the synthetic data enables the creation of a cost-effective, large-scale virtual panel of respondents that is both ``wide" (each respondent has over 50k modeled features) and ``long" (enough samples to reflect a national dataset). However, running LLM inference on the entire U.S. population would be prohibitively expensive, so we employ a sampling strategy. Given the pivotal role of swing states in determining election outcomes, we focus on simulating voter behavior in these states while including representative samples from red and blue states for comparison.

\noindent \textbf{Synthetic Data Generation:}  
The synthetic data used here is generated using the \texttt{SynC} framework \cite{li2020sync}, which reconstructs individual-level data from aggregated sources where collecting real-world individual data is impractical due to privacy, time, or financial constraints. 
\texttt{SynC} is widely recognized and applied across multiple fields to support research and overcome data limitations. For instance, it has been used in outlier detection \cite{Li2020COPOD}, finance \cite{Potluru2023SyntheticFinance}, tabular data modeling \cite{Borisov2022DeepTabularSurvey}, healthcare \cite{Sichani2024SyntheticHealth}, and tourism \cite{merinov2023behaviour}, demonstrating its effectiveness and importance in various domains.

\texttt{SynC} leverages publicly available data, such as the 2023 American Community Survey (ACS), which provides data on 242,338 census block groups, including population statistics and response proportions for each block. 
Using \textit{Data Downscaling}, \texttt{SynC} probabilistically recreates the 340 million residents represented in the aggregated census data.
For our simulation, the synthetic population includes variables relevant to election predictions: (1) age, (2) gender, (3) ethnicity, (4) marital status, (5) household size, (6) presence of children, (7) education level, (8) occupation, (9) individual income, (10) family income, and (11) place of residence.

\texttt{SynC} addresses the challenge of reconstructing individual data $\{x_{m,1}^d, \ldots, x_{m,n_m}^d\}$ from aggregated observations $X_m^d = \sum_{k=1}^{n_m} x_{m,k}^d / n_m$, where $X^d$ is the $d$-th survey question of interest, $m$ is the census block id and $n$ is the number of individuals in $m$. A \textit{Gaussian copula} is employed to model dependencies between survey questions. Given a $d \times d$ covariance matrix $\Sigma$ of the $d$ sruvey questions, the synthetic individuals are drawn as:

\begin{small}
    \begin{equation}
        Z_m^d \sim N(0, \Sigma), \quad u_m^d = \Phi(Z_m^d), \quad X_m^{d} = F^{-1}_d(u_m^d),
    \end{equation}
\end{small}

where $Z_m^d \sim N(0, \Sigma)$ denotes a random seed from a multivariate normal distribution, $\Phi$ is the cumulative distribution function (CDF) of the standard normal distribution, and $F^{-1}_d$ is the inverse CDF of the marginal distribution for feature $d$, which is estimated based on census block level data.
To maintain alignment with aggregated data, SynC uses \textit{marginal scaling}. For categorical variables, it applies a multinomial distribution:

\begin{equation}
    X^d \sim \text{Multi}(1, c^d, p_{m,k}^d),
\end{equation}

where $p_{m,k}^d$ is the probability distribution over $c^d$ categories for question $d$ and individual $k$. Marginal constraints are adjusted iteratively if discrepancies arise between sampled and target proportions.

The multi-phase \texttt{SynC} framework ensures that: (1) marginal distributions of individual features align with real-world expectations, (2) feature correlations are consistent with aggregated data, and (3) aggregated results match the input data. For further details on \texttt{SynC}’s methodology and algorithms, please see the original paper \cite{li2020sync}.

\noindent \textbf{Partition Design and State Categorization:}  
The synthetic dataset evaluation will operate at the state level, where we sample synthetic individuals from each state to simulate voter behavior and aggregate their votes to compare the simulated outcomes with actual election results. 
Given the critical role of swing states and tipping-point states in determining election outcomes, our primary focus is on these states, which include Florida (FL), Wisconsin (WI), Michigan (MI), Nevada (NV), North Carolina (NC), Pennsylvania (PA), Georgia (GA), Texas (TX), Minnesota (MN), Arizona (AZ), and New Hampshire (NH). 
For broader comparison in the following evaluations, we also sample from several reliably ``red states,'' such as Alabama (AL), Arkansas (AR), Idaho (ID), Ohio (OH), and South Carolina (SC), as well as from ``blue states,'' such as California (CA), Illinois (IL), New York (NY), New Jersey (NJ), and Washington (WA). 
These classifications are based on the 2020 election results as described by Wikipedia \cite{wikipedia_swing_state}.

\noindent \textbf{Sampling Method:}  
Running LLM inference on the entire synthetic population is computationally prohibitive, so we adopt a random sampling approach. Each state serves as a sampling unit, with sample sizes ranging between 1/100 and 1/2000 of the synthetic population, depending on the state’s population size. For example, a 1/2000 sampling ratio is applied to highly populated states like California, while a 1/100 ratio is used for smaller states such as New Hampshire. This approach ensures a minimum sample size of $4269$ individuals per state, corresponding to a 1.5\% margin of error at a 95\% confidence level, to maintain sufficient representation. Although our primary focus is on swing states due to their critical influence on election outcomes, we apply the same sampling method to red and blue states included in our simulations to ensure consistency across the analysis.

\subsection{Detailed Evaluation Metrics}
\label{appx.metrics}

To comprehensively evaluate our proposed approaches, we employ multiple metrics for both benchmark datasets (ANES 2016 and 2020 \cite{ANES2016, ANES2020}) and state-level simulations. For the ANES benchmarks, we follow the methodology of \citet{Argyle_Busby_Fulda_Gubler_Rytting_Wingate_2023}, comparing the average voting probabilities:

\paragraph{1. Predicted Proportion ($P(s)$)}
\begin{small}
    \begin{equation}
    \text{Probability} = \frac{\text{Republican Votes}}{\text{Republican Votes} + \text{Democratic Votes}}
    \label{eq:prob_appx}
    \end{equation}
\end{small}

For state-level comparisons, we introduce the following additional metrics:

\paragraph{2. Weighted Absolute Error (WAE)}
\begin{equation}
\label{WAE}
\text{WAE} = \frac{\sum_{s \in S} E(s) \cdot |P(s) - R(s)|}{\sum_{s \in S} E(s)}
\end{equation}
where:
\begin{itemize}
    \item $P(s)$: The simulated proportion, calculated as the ratio of Republican votes to total votes (Republican + Democrat) for each state (\ref{eq:prob_appx}).
    \item $R(s)$: The actual proportion of votes in state $s$.
    \item $E(s)$: The electoral votes assigned to state $s$, serving as weights.
    \item $S$: The set of all selected states.
\end{itemize}

\paragraph{3. Weighted Mean Squared Error (WMSE)}
\begin{equation}
\label{WMSE}
\text{WMSE} = \frac{\sum_{s \in S} E(s) \cdot (P(s) - R(s))^2}{\sum_{s \in S} E(s)}
\end{equation}
where:
\begin{itemize}
    \item $(P(s) - R(s))^2$: The squared error between the simulated and actual proportions for each state.
\end{itemize}

\paragraph{4. Bias Metric (BM)}
\begin{equation}
\label{BM}
\text{BM} = \frac{\sum_{s \in S} E(s) \cdot (P(s) - R(s))}{\sum_{s \in S} E(s)}
\end{equation}
where:
\begin{itemize}
    \item \textbf{Positive Value}: Reflects a systematic overestimation of $P(s)$, indicating a bias toward the Republican Party.
    \item \textbf{Negative Value}: Reflects a systematic underestimation of $P(s)$, indicating a bias toward the Democratic Party.
\end{itemize}

These metrics are calculated across the entire sample to evaluate both the magnitude and direction of errors. Accuracy is further assessed by comparing the predicted winning party with the actual election outcome.

For the synthetic dataset, we treat each state as an independent validation unit. The simulated results—both in terms of the winning candidate and vote share percentages—are compared against the actual 2020 election results for each state. Accuracy is evaluated based on:
\begin{enumerate}
    \item Agreement between the predicted and actual winning candidate for each state.
    \item Aggregate performance across all states, ensuring the model captures overall election trends.
\end{enumerate}

This state-level evaluation leverages voter-level information processed through LLMs to generate accurate simulations, providing a robust assessment of model performance across diverse electoral scenarios.

\section{Evaluations on Synthetic Personas for the 2020 U.S. Population} 

\label{appx:validation}

\begin{figure}[!ht]
    \centering
    \includegraphics[width=\linewidth]{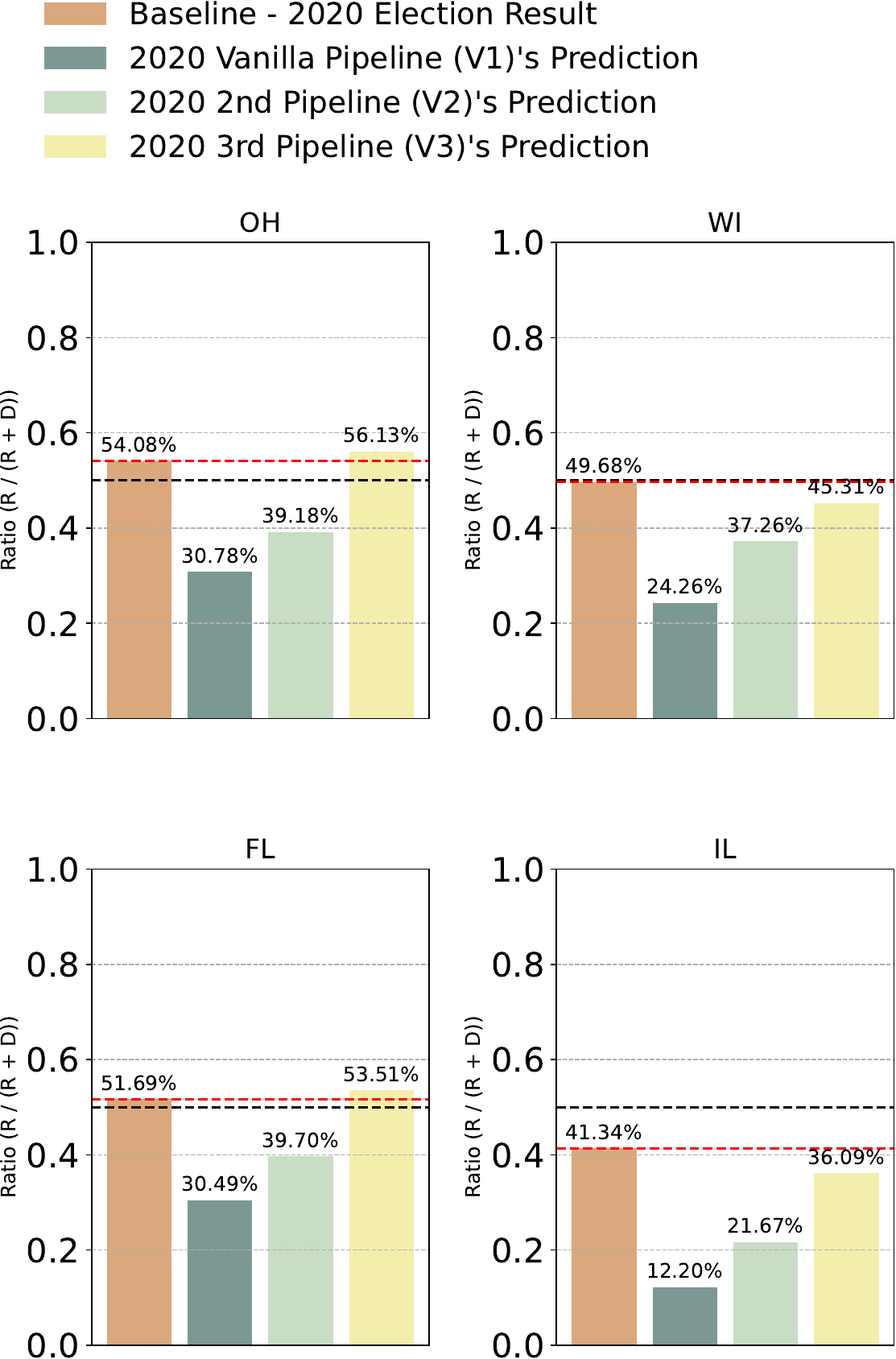}  
    \caption{LLM’s simulations for four states in the 2020 election compared with Ground Truth results. The figure presents results for one red state (Ohio, OH), one blue state (Illinois, IL), one swing state (Wisconsin, WI), and one tipping-point state (Florida, FL). 
    V1 and V2 pipelines tend to underestimate Republican support, while V3 (Multi-step Reasoning) provides the closest alignment with actual outcomes, especially in swing and tipping-point states.
    }
    \label{fig:2020example}  
    \vspace{-1em}  
\end{figure}

In addition to the nationwide evaluation on the ANES datasets, we conducted state-level simulations using synthetic data to compare it with actual 2020 election outcomes. 
For each state, we performed random sampling based on population size to ensure a statistically meaningful number of personas. The simulation outcomes were then benchmarked against official 2020 Presidential General Election Results from the Federal Election Commission (FEC). As in the benchmark evaluations, we calculated the average voting probabilities to assess the alignment of predictions with real-world outcomes.
We evaluated five red states, five blue states, and 11 swing and tipping-point states. Figure~\ref{fig:2020example} highlights representative results from these categories, providing insights into the model’s performance in different electoral contexts.

\begin{figure}[!t]
    \centering
    \includegraphics[width=\linewidth]{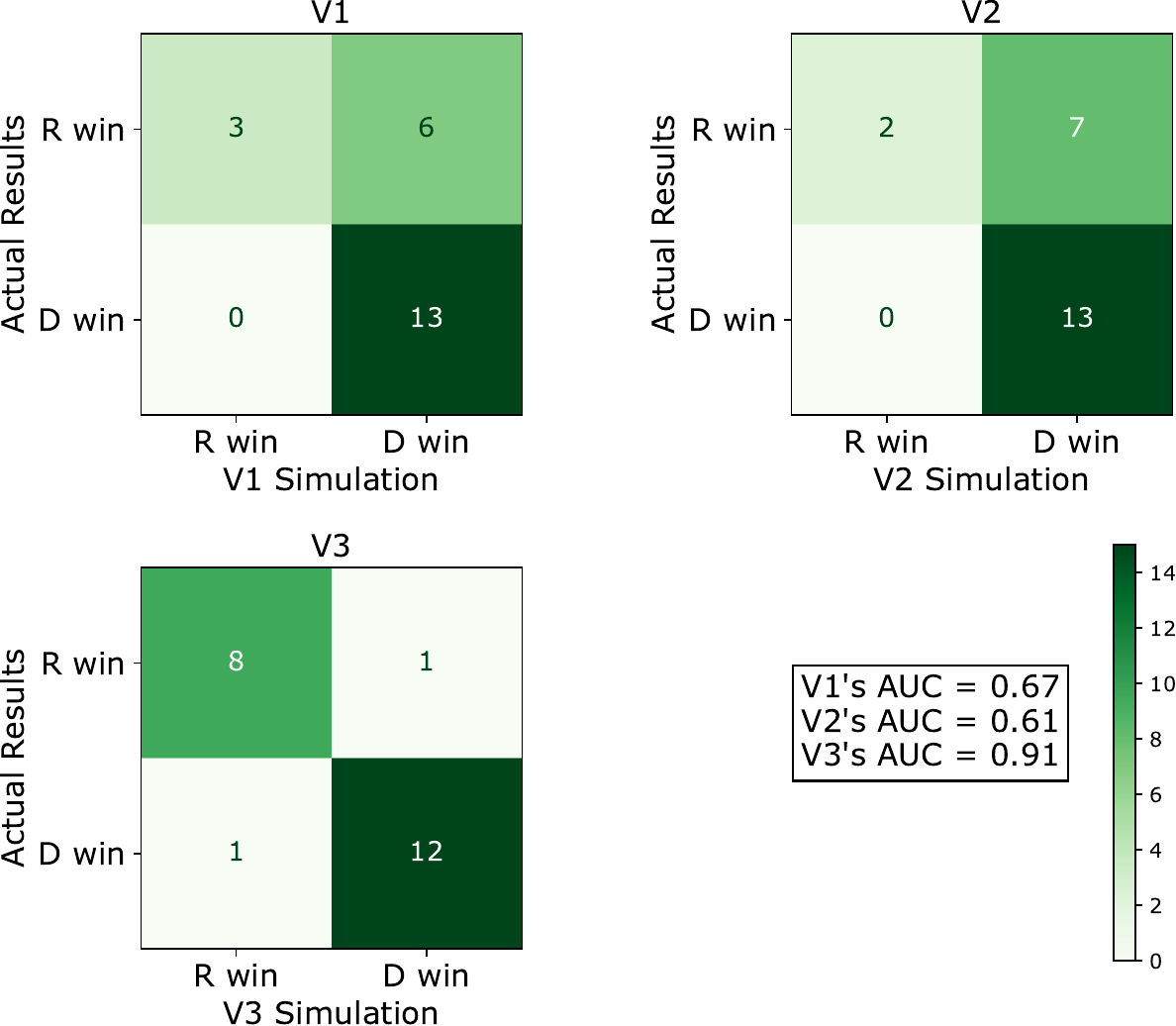}  
    \caption{
    Aggregated results of the three pipelines (V1, V2, V3) on state-level simulations. Each confusion matrix presents the number of states where predictions align with or deviate from actual outcomes. V1 (AUC = 0.69) and V2 (AUC = 0.62) show lower accuracy, while V3 (AUC = 0.90) performs best, effectively capturing Republican victories without compromising Democratic predictions. It is worth noting that, so far, we have only tested the pipelines in 21 states. If the scope is expanded to include all states, the AUC of V3 is expected to improve further, while the AUC of V1 and V2 are expected to decline.
    }
    \vspace{-0.2in}
    \label{fig:2020aggregate}  
\end{figure}

Consistent with the ANES dataset evaluations, the V1 pipeline (Demographic-only Prompt) exhibited a skew toward the Democratic Party, even in traditionally Republican-leaning states like South Carolina (SC), Alabama (AL), and Ohio (OH), with predictions diverging significantly from actual results. This illustrates the limitations of using demographic data alone without time-sensitive context.
The V2 pipeline (Time-dependent Prompt) introduced election-year-specific information, which partially reduced the skew in the state-level simulations. However, the model still struggled to eliminate prediction biases, particularly in polarized states. Interestingly, this differed from the ANES evaluations, where including time-dependent information amplified the bias.
The V3 pipeline (Multi-step Reasoning) demonstrated the most accurate performance, effectively mitigating skewness across deep red and blue states. In these polarized states, the predictions closely mirrored the actual voting outcomes, reflecting the model’s improved ability to incorporate ideological alignment through multi-step reasoning.

For swing and tipping-point states, the V3 pipeline achieved robust results, correctly simulating the outcomes in 9 out of 11 states. Minor deviations were observed in North Carolina (NC) and Arizona (AZ), where the predictions were slightly misaligned with the real results. Nonetheless, the V3 pipeline provided balanced predictions that accurately captured the competitive dynamics typical of swing states, further validating its effectiveness.

In summary, the comparative performance of the three pipelines across different state categories is shown in Figure~\ref{fig:2020example}. The V3 pipeline consistently outperformed the other two, delivering more stable and accurate predictions. Aggregate results for all pipelines on all 21 chosen states is shown in the below figure ~\ref{fig:2020aggregate}.

\section{Additional Results on 2024 Prediction}
\label{appx:2024results}

\begin{figure*}[!t]
    \centering
    \includegraphics[width=\textwidth]{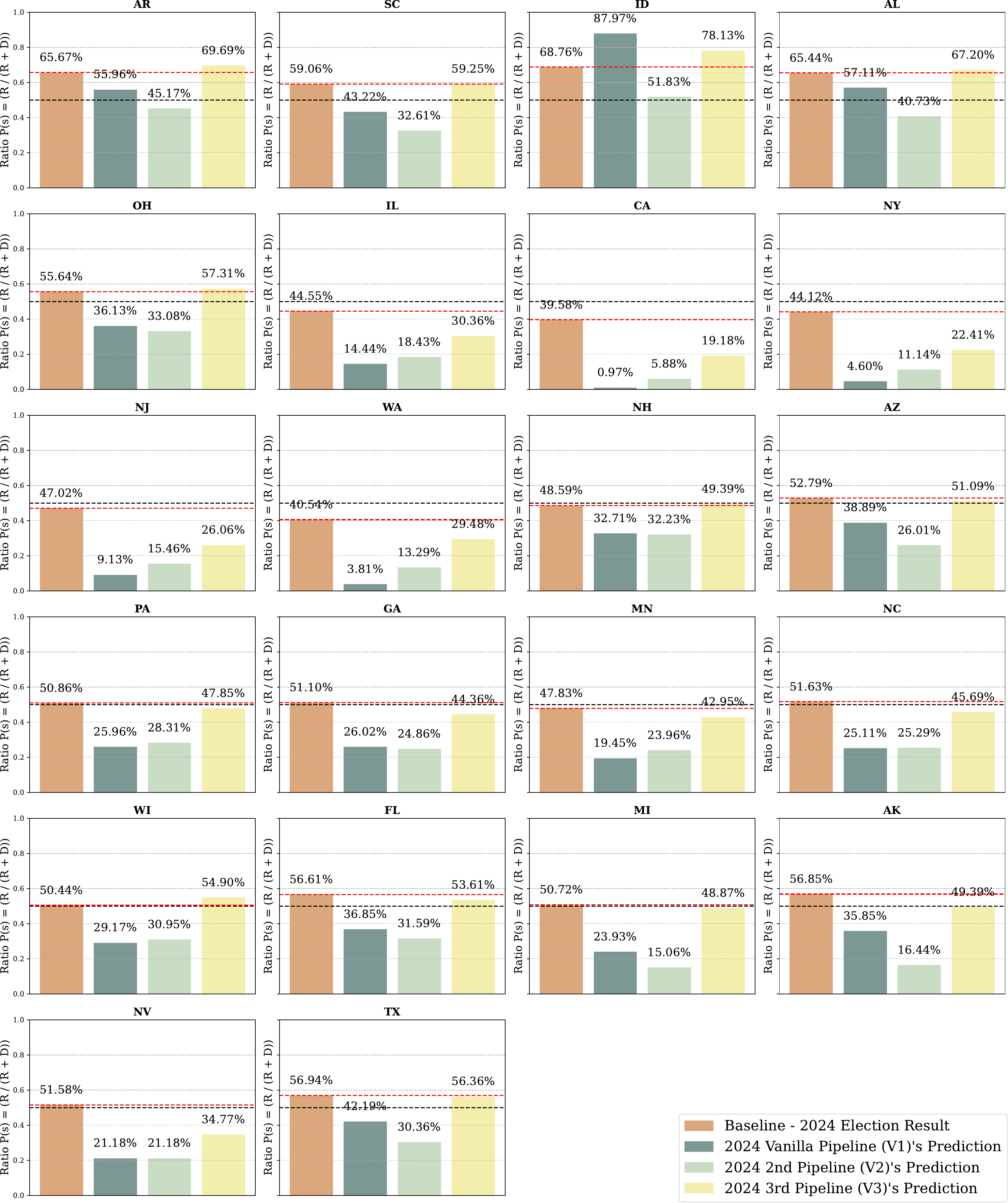}  
    \caption{
    Overall performance of the three pipelines (V1, V2, V3) in the 2024 simulation across five red states (Arkansas (AR), South Carolina (SC), Idaho (ID), Alabama (AL), Ohio (OH)), five blue states (Illinois (IL), California (CA), New York (NY), New Jersey (NJ), Washington (WA)), 11 swing and tipping-point states (New Hampshire (NH), Arizona (AZ), Pennsylvania (PA), Georgia (GA), Minnesota (MN), North Carolina (NC), Wisconsin (WI), Florida (FL), Michigan (MI), Nevada (NV), Texas (TX)), and an additional red state (Alaska (AK)). The red reference line corresponds to the 2024 election results reported by the NBC News \cite{nbc2024election}, while the black reference line represents an equal vote share (0.5) between the two parties.}
    \label{fig:2024results}  
    \vspace{1em}  
\end{figure*}

The 2024 state-level Simulations offer deeper insights into the performance of the proposed pipelines across diverse electoral contexts. As discussed in \S \ref{subsec:study1}, simulations for the 2024 election indicate a systematic bias toward the Democratic Party across the 11 swing and tipping-point states. This bias may reflect the LLM’s sensitivity to candidate-specific factors in the 2024 context. Specifically, prior to the election, V3 (\S \ref{subsec:v3}) was evaluated across all 50 states, while V1 (\S \ref{subsec:v1}) and V2 (\S \ref{subsec:v2}) were tested on selected swing states and traditional red and blue states. The comparative performance of these pipelines is presented in Figure~\ref{fig:2024results}.

At the state level, several notable shifts are observed compared to 2020 predictions. For instance, Wisconsin (WI) demonstrates a significant change, with Trump projected to win 54.90\% of the vote. Gains are also observed in Pennsylvania (PA) 47.85\%, Michigan (MI) 48.87\%, and New Hampshire (NH) 49.49\%, though these states remain highly competitive. 

In other key battleground states, Arizona (AZ) is forecasted to return to the Republicans with 51.09\%, while Florida (FL) and Texas (TX) continue to show strong Republican support at 53.62\% and 56.36\%, respectively. Conversely, in contrast to the actual results, Nevada (NV) at 34.77\%, Georgia (GA) at 44.36\%, and Minnesota (MN) at 42.95\% are predicted to lean more toward the Democratic Party, highlighting the complex dynamics of these closely contested regions.

In traditional strongholds, the predictions align with historical trends. Republican-dominated states like Arkansas (AR) and Alabama (AL) continue to show robust GOP support, while Democratic bastions such as California (CA), New York (NY), and Illinois (IL) remain reliably blue. An exception is Alaska (AK) 49.39\%, where the model predicts a closer contest compared to prior elections.

These state-level results, summarized in Figure~\ref{fig:2024results}, highlight the nuanced performance of the pipelines. The varying prediction patterns underscore both the strengths and limitations of the models, emphasizing opportunities for further refinement to better capture the complexities of voter behavior and electoral dynamics.

\section{Beyond Accuracy}
\label{E}
\begin{figure}[!h]
    \centering
        \centering
        \includegraphics[width=\linewidth]{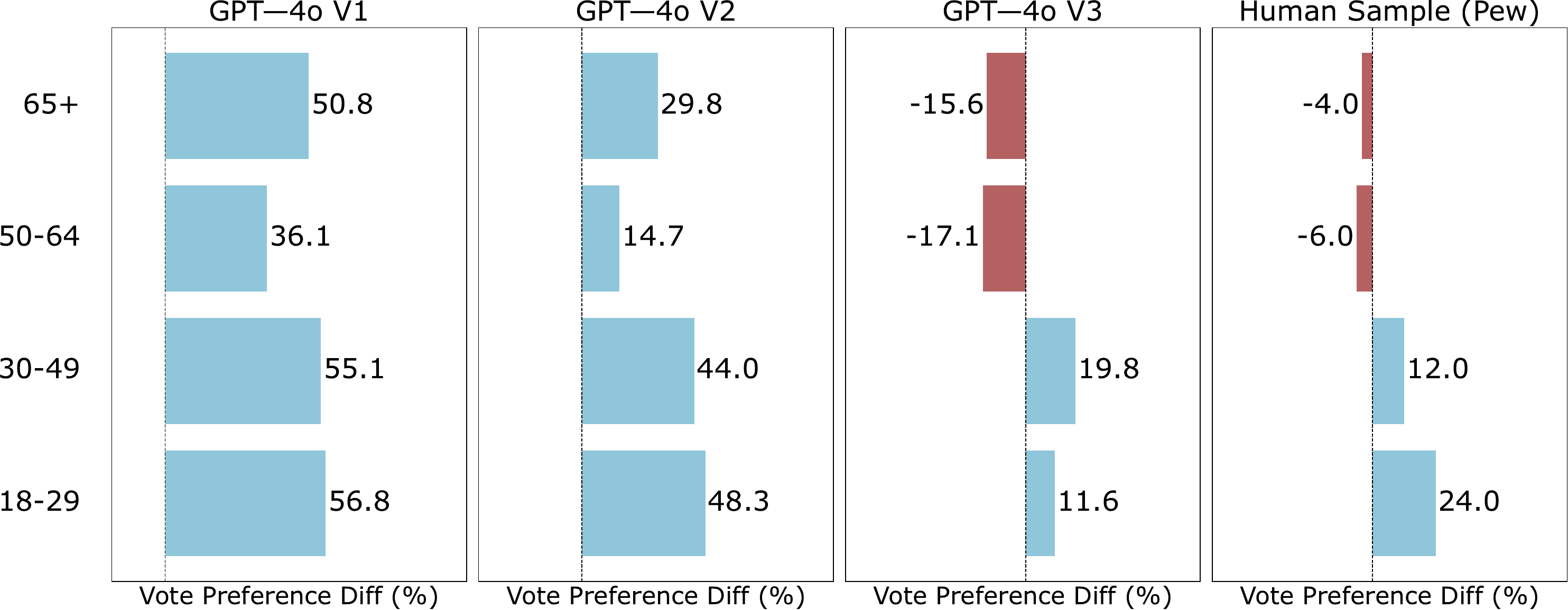}
        \caption{Demographic (Age) voting pattern: comparing LLM Simulations with real human data (Pew Report)}
        \label{fig:age_gap}
    \vspace{-0.1in}
\end{figure}

\begin{figure}[!ht]
    \centering
        \centering
        \includegraphics[width=\linewidth]{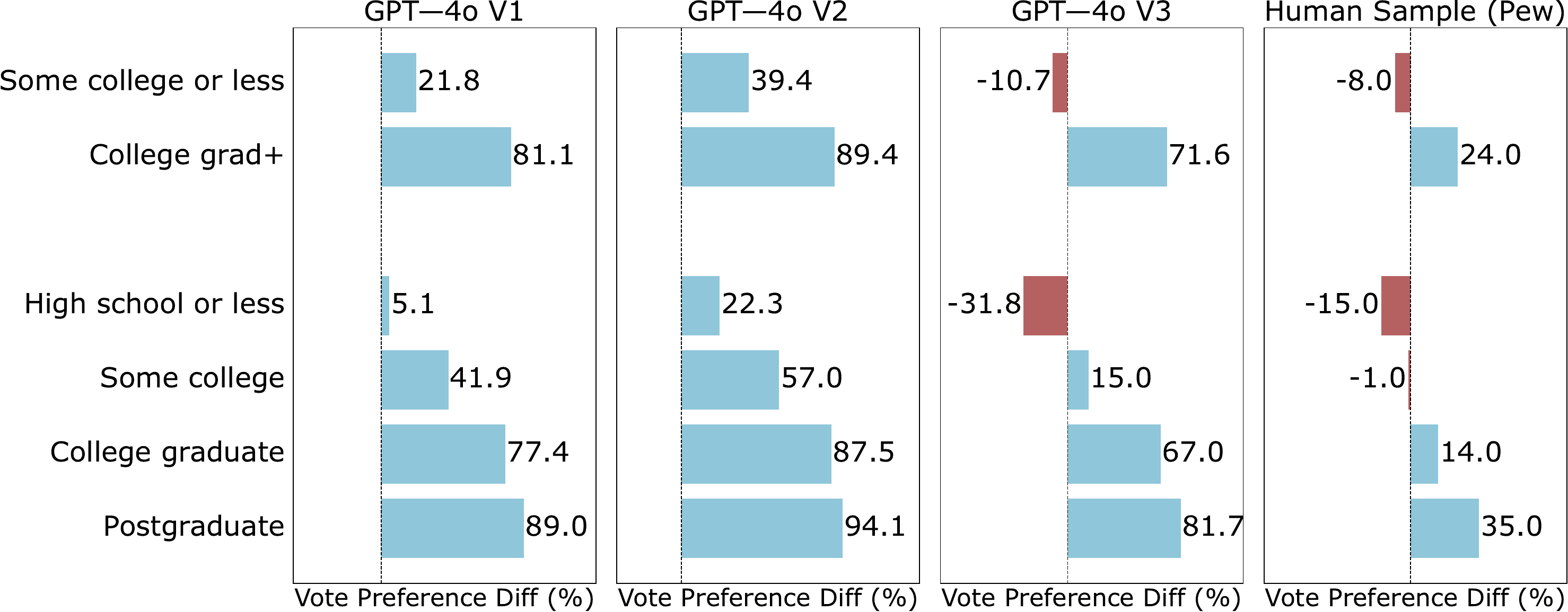}
        \caption{Demographic (Education) voting pattern: comparing LLM Simulations with real human data (Pew Report)}
        \label{fig:education_gap}
    \vspace{-0.1in}
\end{figure}

\section{Broader Impact Statement}
\vspace{-0.5ex}
This work explores the application of LLMs to simulate voter behaviour through enhanced reasoning and data synthesis, which may inform policymakers, researchers, and journalists, aiding them in understanding voter behavior and electoral outcomes. 
By offering a more transparent and adaptable approach to prediction, this research may help demystify complex political processes, reduce reliance on narrow historical data, and guide strategic resource allocation for stakeholders.

\begin{figure*}[!t]
    \centering
    \includegraphics[width=\textwidth]{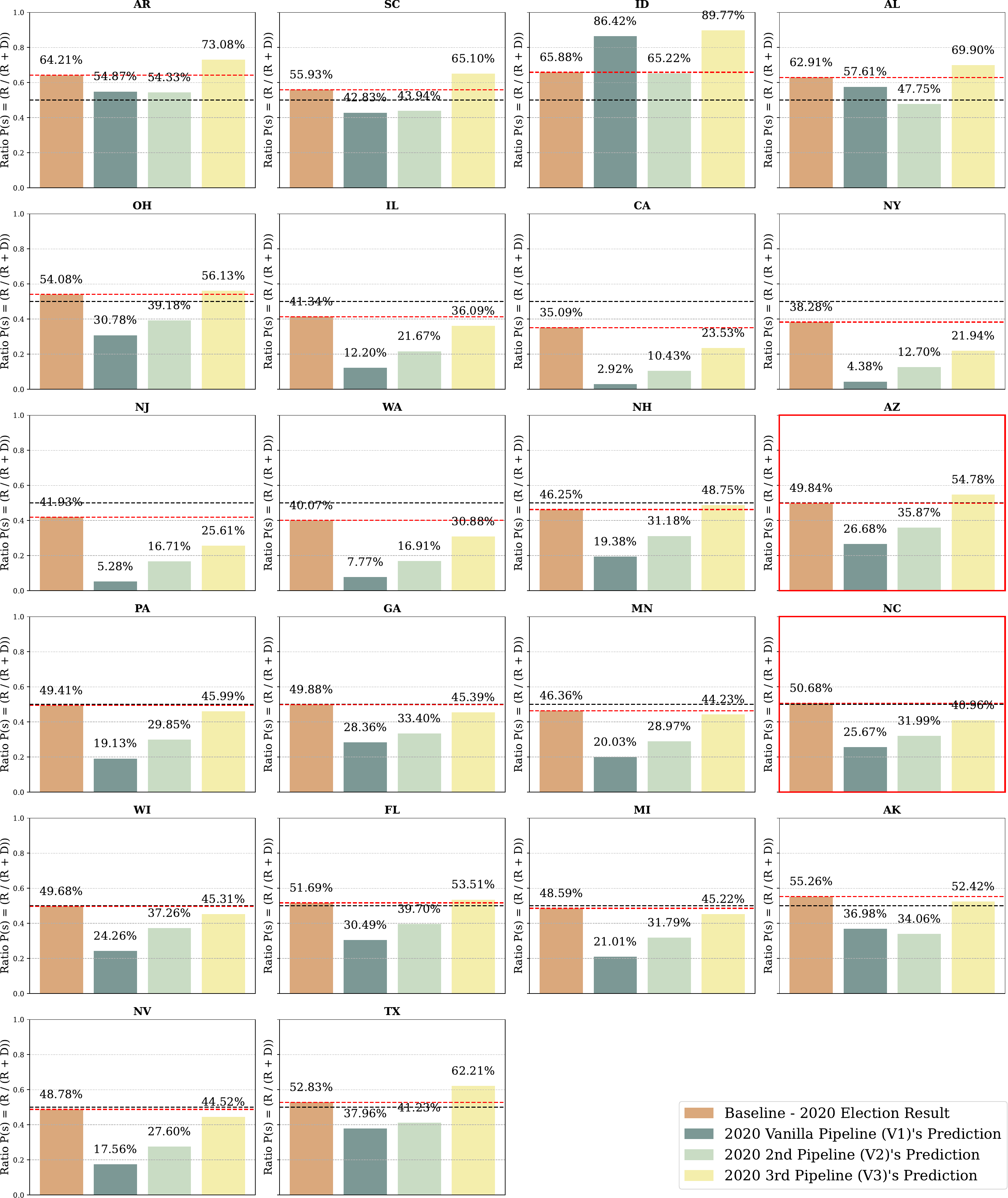}  
    \caption{
    Overall performance of the three pipelines (V1, V2, V3) in the 2020 simulation across five red states (Arkansas (AR), South Carolina (SC), Idaho (ID), Alabama (AL), Ohio (OH)), five blue states (Illinois (IL), California (CA), New York (NY), New Jersey (NJ), Washington (WA)), 11 swing and tipping-point states (New Hampshire (NH), Arizona (AZ), Pennsylvania (PA), Georgia (GA), Minnesota (MN), North Carolina (NC), Wisconsin (WI), Florida (FL), Michigan (MI), Nevada (NV), Texas (TX)), and an additional red state (Alaska (AK)). The red reference line corresponds to the 2020 election results \cite{federal2020elections}, while the black reference line represents an equal vote share (0.5) between the two parties.}
    \label{fig:2020results}  
    \vspace{1em}  
\end{figure*}

\end{document}